\documentclass[conference]{IEEEtran}
\IEEEoverridecommandlockouts
\usepackage{cite}
\usepackage{amsmath,amssymb,amsfonts}
\usepackage{graphicx}
\usepackage{textcomp}
\usepackage{xcolor}
\def\BibTeX{{\rm B\kern-.05em{\sc i\kern-.025em b}\kern-.08em
    T\kern-.1667em\lower.7ex\hbox{E}\kern-.125emX}}

\usepackage[utf8]{inputenc} 
\usepackage[T1]{fontenc}    
\usepackage{hyperref}       
\usepackage{url}            
\usepackage{booktabs}       
\usepackage{amsfonts}       
\usepackage{nicefrac}       
\usepackage{microtype}      
\usepackage{lipsum}		
\usepackage{graphicx}
\usepackage{subcaption}
\usepackage{graphicx}
\usepackage{xcolor} 
\usepackage{tikz}
\usetikzlibrary{shapes.geometric, arrows, positioning, automata,positioning,fit,backgrounds}
\tikzstyle{process} = [rectangle, minimum width=2cm, minimum height=1cm, text centered, draw=black, fill=white!30]
\tikzstyle{sum} = \tikzstyle{sum} = [draw, circle, minimum size=.5cm]
\tikzstyle{arrow} = [thick,->,>=stealth]

\usepackage{amsmath}

\usepackage{bm}
\usepackage{units}
\usepackage{float}
\usepackage{dblfloatfix}
\usepackage{algorithm}
\usepackage[noend]{algpseudocode}
\makeatother
\usepackage[T1]{fontenc}

\usepackage{todonotes}

\begin{document}

\title{A comparison of controller architectures and learning mechanisms for arbitrary robot morphologies}

\author{
\IEEEauthorblockN{Jie Luo\textsuperscript{1}, Jakub Tomczak\textsuperscript{2}, Karine Miras\textsuperscript{1}, and Agoston E. Eiben\textsuperscript{1}}
\IEEEauthorblockA{
\textit{Computer Science Dept.}\\
\textsuperscript{1}Vrije Universiteit Amsterdam, Netherlands\\
\textsuperscript{2}Eindhoven University of Technology, Netherlands\\
Email: j2.luo@vu.nl}
}


\maketitle

\begin{abstract}
The main question this paper addresses is: What combination of a robot controller and a learning method should be used, if the morphology of the learning robot is not known in advance? Our interest is rooted in the context of morphologically evolving modular robots, but the question is also relevant in general, for system designers interested in widely applicable solutions. We perform an experimental comparison of three controller-and-learner combinations: one approach where controllers are based on modelling animal locomotion (Central Pattern Generators, CPG) and the learner is an evolutionary algorithm, a completely different method using Reinforcement Learning (RL) with a neural network controller architecture, and a combination `in-between' where controllers are neural networks and the learner is an evolutionary algorithm. We apply these three combinations to a test suite of modular robots and compare their efficacy, efficiency, and robustness. Surprisingly, the usual CPG-based and RL-based options are outperformed by the in-between combination that is more robust and efficient than the other two setups. 
\end{abstract}


\section{Introduction}
Enabling robots to learn tasks automatically is an important feature on its own, and also necessary within an evolutionary robot system, where both the morphologies (bodies) and the controllers (brains) are developed by evolution. In such systems, `newborn' robots should undergo a learning phase to fine-tune the inherited brain to the inherited body quickly after birth \cite{Eiben2013, Eiben2020}. This raises the question: what combination of a robot controller and a learning method should be used in the robots' morphology which is not known in advance? In general, a robot's ability to learn a task depends on three major system components, namely, the body (morphology, hardware), the brain (controller, software), and the learning algorithm. 

In the current literature, the majority of studies investigate controller optimization using multiple learning algorithms, but focusing on a specific control architecture \cite{Diggelen2021a, LeGoff2020}; comparisons of different control architectures and learning methods for learnable controllers and arbitrary modular robots are rarely carried out.

This study makes a step towards closing this gap by comparing three different combinations of a specific control architecture and a learning algorithm. The possible control architectures are Central Pattern Generator (CPG), Artificial Neural Network (ANN) and Deep Reinforcement Learning (DRL) policy controller. The possible learning algorithms are Reversible Differential Evolution (RevDE) \cite{weglarz2021population} representing semi-supervised learning and Proximal Policy Optimization (PPO) \cite{schulman2017proximal} representing reinforcement learning. The combinations we compare here are CPG+RevDE, DRL+PPO, and ANN+RevDE. The motivation behind these choices is as follows. Using CPGs is a well-established, biologically plausible option to control modular robots actuated through joints, where learning can be performed by any heuristic black-box optimization method. RevDE is one such method that proved to be successful in the past for this application. Deep Reinforcement Learning is also a straightforward and increasingly popular option for robot learning with implications for the appropriate controller architecture, namely the use of ANNs. Additionally, we test the ANN+RevDE combination as an `in-between' option that, to our best knowledge, has not been investigated previously. 

The main contribution of this work is threefold:
\begin{enumerate}
\item It demonstrates a test-suite based approach to experimental research into robot learning, where the robots that make up the test suite are not only hand-picked, but also generated algorithmically. 
\item Furthermore, the controller-and-learner combinations are not only compared by the usual performance measures, efficiency and efficacy, but also by robustness, i.e., stability or consistency over the different robot morphologies.
\item It provides an empirical assessment of three options, including two `usual suspects' that researchers in the field are likely to consider: CPG-based controllers with a good weight optimizer and a Deep Reinforcement Learning method. The results indicate a surprising outcome, both of these methods are outperformed by the third one, ANN+RevDE. 
\end{enumerate}

\section{Related Work}
\subsection{Robot Controllers}
A popular class of controllers is based on utilizing Artificial Neural Networks (ANN). The optimization of an ANN is typically done by approximating gradients for gradient-based methods or by applying derivative-free methods to alter internal weights and biases of all neurons within the ANN \cite{lee2003evolving,pollack2000golem,pollack2003computer,lipson2006evolving,lund2003co,poikselka2015evolutionary, ranganath2011distributed}. Alternatively, reinforcement learning (RL) could be used to update the controller \cite{bhatia2022,D'Angelo2013, Shen2012}. Here, we focus on one specific implementation of RL that utilizes two networks: a controller network (also called a policy controller), and a surrogate model, an additional neural network -- a critic network -- to update the parameters of the controller.

A popular approach relies on the idea inspired by biology that aims at creating rhythmic patterns to control the motion of the robots. These approaches use different controller types and learning algorithms for creating rhythmic patterns. Early approaches used Control Tables \cite{Bongard2006,Yim2000}, where each column of a table contains a set of actions for a module in the configuration, and Simple Sinusoidal, in which a specific sinusoidal function is utilized for each motor providing an easy way to parameterize a control pattern \cite{gonzalez2006motion,bruce2014design,faina2011first}. These methods were followed by a controller architecture called Cyclic Splines \cite{jelisavcic2017real,kober2009policy} in which a spline is fitted through a set of action points in time to define a periodic control sequence (\textit{i.e.} control policy). Another successful (bio-inspired) controller called Central Pattern Generators (CPGs) \cite{Ijspeert2008} was based on the spinal cord of vertebrates and can produce stable and well-performing gaits on both non-modular robots \cite{Christensen2013, Sproewitz2008} and modular robots \cite{Sproewitz2008, Pouya2010, Luo2022}. CPGs are biological neural circuits that produce rhythmic output in the absence of rhythmic input \cite{Bucher2015}. In this work, we use CPGs to parameterize a controller and create biologically plausible motion patterns. 

\subsection{Controller Learning Algorithms}
The problem of controller learning in robotics could be phrased as the \textit{black-box optimization} problem \cite{audet2017derivative, jones1998efficient} since we need to either run a simulation or a physical robot to obtain a value of the objective function (or the fitness function). There is a vast amount of literature on learning algorithms on only one type of controller \cite{Weel2017, Diggelen2021, Lan2020, LeGoff2020}, naming only a few.

In \cite{Diggelen2021}, a comparison of three learning algorithms in modular robots is performed where \textit{NIP-Evolutionary Strategies}, \textit{Bayesian Optimization} and \textit{Reversible Differential Evolution} (RevDE) \cite{Tomczak2020} are tested. The outcome of this study indicates that the shape of the fitness landscape in evolutionary strategies hints at a possible bias for morphologies with many joints. This could be an unwanted property for the implementation of lifetime learning because an algorithm should work consistently on different kinds of morphologies. Bayesian Optimization is good at sample efficiency, however, it requires much more time compared to the other two methods due to the higher time complexity (cubic complexity). The best-performing algorithm in this comparison was RevDE which scales well in terms of complexity and generalizes well across various morphologies. Therefore, we use RevDE in this paper. Moreover, we apply Proximal Policy Optimization (PPO) in the context of RL. PPO is a family of model-free RL learning algorithms that search the space of policies rather than assigning values to state-action pairs \cite{franccois2018introduction}. It was used in recent research \cite{Gupta2021} and performs well across various morphologies.

\section{Methodology}


\subsection{Robot Controllers}
In this research, our task is gait learning, therefore the controllers we use are open-loop controllers without steering. The choice of a robot controller is a crucial design decision and determines the resulting search space and, as a consequence, the behaviour of a robot. Different types of controllers may require different inputs, e.g. DRL-Policy controller and ANN controller need observations from the environment as input, however, CPG does not reply on observation in an open-loop controller. Moreover, the number of parameters to be optimized in each type of controller can be different. Last but not least, the outputs differ too. CPG and ANN controllers output actions to the hinges directly while the DRL-Policy controller output the action distribution.

\subsubsection{CPG controller}
Each robot hinge i is associated with a CPG that is defined by three neurons: a $x_i$-neuron, a $y_i$-neuron, and an $out_i$-neuron, which are recursively connected to produce oscillatory behaviour. 

The CPG network structure we used has two layers:
\begin{enumerate}
    \item Internal connection: The change of the $x_i$ and $y_i$ neurons' states with respect to time is calculated by multiplying the activation value of the opposite neuron with a weight. To reduce the search space, we define $w_{x_iy_i}$ to be $-w_{y_ix_i}$ and call their absolute value $w_i$ and set $w_{x_io_i}$ =1. The resulting activations of neurons $x$ and $y$ are periodic and bounded. The initial states of all $x$ and $y$ neurons are set to $\frac{\sqrt{2}}{2}$ because this leads to a sine wave with amplitude 1, which matches the limited rotating angle of the joints.

    \item External connection: CPG connections between neighbouring hinges. Two hinges are said to be neighbours if their tree-based distance (how many edges between one node and the other) is less than or equal to two. 

     $x$ neurons depend on neighbouring $x$ neurons in the same way as they depend on their $y$ partner. Let $i$ be the number of the hinge, N\textsubscript{i} the set of indices of hinges neighbouring hinge $i$, and $w_{ij}$ the weight between $x_i$ and $x_j$. Again, $w_{ji}$ is set to be $-w_{ij}$. The extended system of differential equations is then:
    \begin{align}\label{eq:ODE_CPG}
         \begin{split}
            &\Dot{x}_i = w_iy_i + \sum_{j \in \mathcal{N}_i} w_{x_jx_i}x_{j} \\
            &\Dot{y}_i = w_ix_i
         \end{split}
    \end{align}

Because of this addition, $x$ neurons are no longer bounded between $[-1,1]$. To achieve this binding, we use a variant of the sigmoid function, the hyperbolic tangent function (tanh), as the activation function of $out_i$-neurons.
\end{enumerate}

The total number of the weights parameters per robot we have to optimise for the CPG network is the sum of weights of these two connections: CPG\_N\textsubscript{param}= N\textsubscript{hinges} + N\textsubscript{i}
Take the spider for example, it has 8 CPGs (hinges) and 10 pairs of neighbouring connections between CPGs, therefore the total number of the weights parameters is 18. 

\subsubsection{ANN controller}
In an ANN robot controller, the ANN internally connects an input layer of neurons to an output layer that triggers the actuators, possibly via a layer of hidden neurons. The output of a previous layer is multiplied by corresponding weights before being summed with a bias term and thus serves as input for the next layer. 
Here, the main components of the ANN (a.k.a. Actor network) are:

\begin{enumerate}

    \item \textit{Single Observation Encoders}. A sub-network for encoding a single type of observation. In our research, we use two types of observations: state of each hinge (activation of the hinge between -1 and 1 which is its motion range) and the orientation of the robot (based on the core modular of the robot). The input of the coordinates observation network is $N\textsubscript{hinges}\cdot3$ dimensions and the input of the orientation observation network which is 4 dimensions. The output of both networks is a 32-dimensional vector through a linear layer followed by a tanh activation function.

    \item \textit{Observation Encoder} A network that concatenates the encoded observations. It receives inputs from the two Single Observation Encoders and passes the encoded observations which are [32+32=64] dimensional through a linear layer followed by a tanh activation function to produce the final output of a 32-dimensional vector.

    \item \textit{Actor} Takes the concatenated encoded observations as input and outputs the action to be taken by the robot. The dimension of the action is based on the number of the robot's hinges. 
     
\end{enumerate}

The total number of parameters per robot to be optimized is equal to the sum of the parameters of the Single Observation Encoder, Observation Encoder, and Actor: $ANN\_N\textsubscript{param}=32\cdot(N\textsubscript{hinges} \cdot 3+4+1)+32\cdot(64+1)+ N\textsubscript{hinges}\cdot(32+1)$.

\subsubsection{DRL-Policy controller}
The Deep Reinforcement Learning (DRL) paradigm provides a way to learn efficient representations of the environment from high-dimensional sensory inputs, and use these representations to interact with the environment in a meaningful way. At each time-step, the robot senses the world by receiving observations $o_t$ provided by the simulator, then it takes an action $a_t$, and is given a reward $r_t$. A policy $\pi_\theta$($a_t$ $\vert$ $o_t$) models the conditional distribution over action $a_t$ $\in$ A given an observation $o_t$ $\in$ O($s_t$). The goal is to find a policy which maximizes the expected cumulative reward R under a discount factor $\gamma \in (0, 1)$. 

\textbf{Policy controller}
Policy $\pi_\theta$, as the robot’s behaviour function, tells us which action to take in state s. In our research, the implementation of the policy controller has an Actor network, a Critic network, and an ActorCritic network that merges the two. 
\begin{enumerate}
    \item Actor network: a deep neuron network that outputs a Gaussian distribution over the possible actions given an observation. Similar to the ANN controller, observations are encoded into a 32-dimensional vector, but instead of producing actions directly, it produces the action probability using two hidden layers (mean\_layer and std\_layer).

    \item Critic network: a deep neuron network which outputs a single scalar value that approximates the expected return of the current state of the input observation.

    \item ActorCritic network: the primary component that combines the Actor and Critic networks and allows for sampling actions or computing their probabilities and the value of an observation. It can either output the action distribution, the state-value function or both, along with the log-probability of the actions taken and the entropy of the action distribution.
\end{enumerate}

The robot chooses its action via the policy $\pi_\theta$ where $\theta$ are the parameters of these three NNs which will be optimized by a DRL algorithm called Proximal Policy Optimization (PPO). 

The total number of parameters per robot we have to optimize for ActorCritic Network is equal to the sum of the parameters of the Actor, Critic, and ObservationEncoder sub-modules: $DRL\_N\textsubscript{param}=(N\textsubscript{hinges}\cdot(32+1) + 2\cdot N\textsubscript{hinges}\cdot(N\textsubscript{hinges}+1))+ (1\cdot32+1) + (32\cdot(N\textsubscript{hinges} \cdot 3+4+1)+32\cdot(64+1))$.

\subsection{Learning Methods}

The problem of learning a robot controller is stated as a maximization problem of a function (reward or fitness) that is non-differentiable and could be given only after running a real-world experiment or a simulation. Since we cannot calculate the gradients concerning the controller weights, we must apply other learning methods that either utilize approximate gradients (e.g., through surrogate models) or derivative-free methods. In the following paragraphs, we present details of a specific derivative method (RevDE) and an instance of RL (PPO).



\subsubsection{RevDE}
In a recent study on modular robots \cite{Diggelen2021}, it was demonstrated that Reversible Differential Evolution (RevDE) \cite{weglarz2021population}, an altered version of Differential Evolution, performs and generalizes well across various morphologies. This method works as follows \cite{Tomczak2020}:
\begin{enumerate}
    \item Initialize a population with \textit{$\mu$} samples ($n$-dimensional vectors), $\mathcal{P}_{\mu}$. 
    \item Evaluate all \textit{$\mu$} samples.
    \item Apply the reversible differential mutation operator and the uniform crossover operator.\\
    \textit{The reversible differential mutation operator}: Three new candidates are generated by randomly picking a triplet from the population, $(\mathbf{w}_i,\mathbf{w}_j,\mathbf{w}_k)\in \mathcal{P}_{\mu}$, then all three individuals are perturbed by adding a scaled difference in the following manner:
        \begin{equation}\label{eq:de3}
            \begin{split}
            \mathbf{v}_1 &= \mathbf{w}_i + F \cdot (\mathbf{w}_j-\mathbf{w}_k) \\
            \mathbf{v}_2 &= \mathbf{w}_j + F \cdot (\mathbf{w}_k-\mathbf{v}_1) \\
            \mathbf{v}_3 &= \mathbf{w}_k + F\cdot (\mathbf{v}_1-\mathbf{v}_2) 
            \end{split}
        \end{equation}
        where $F\in R_+$ is the scaling factor. New candidates $y_1$ and $y_2$ are used to calculate perturbations using points outside the population. This approach does not follow the typical construction of an EA where only evaluated candidates are mutated.\\
        \textit{The uniform crossover operator}: Following the original DE method \cite{Storn1997}, we first sample a binary mask $\mathbf{m} \in \{0, 1\}^D$ according to the Bernoulli distribution with probability \textit{$CR$} shared across $D$ dimensions, and calculate the final candidate according to the following formula:
        \begin{equation}\label{eq:de2}
              \mathbf{u} = \mathbf{m} \odot \mathbf{w}_n+(1-m) \odot \mathbf{w}_n .
        \end{equation}
        Following general recommendations in literature \cite{Pedersen2010} to obtain stable exploration behaviour, the crossover probability CR is fixed to a value of $0.9$ and the scaling factor $F$ is fixed to a value of 0.5. 
    \item Perform a selection over the population based on the fitness value and select \textit{$\mu$} samples.
    \item Repeat from step (2) until the maximum number of iterations is reached.
\end{enumerate}

As explained above, we apply RevDE here as a learning method for our robot zoo. In particular, it will be used to optimize the weights of the CPGs and the parameters of ANN controllers of our modular robots for the task. 

\subsubsection{PPO}
We use the Proximal Policy Optimization (PPO) \cite{schulman2017proximal} algorithm to optimize a policy. It improves training stability by using a clipped surrogate objective enforcing a divergence constraint on the size of the policy update at each iteration so that the parameter updates will not change the policy too much per step. Let us denote the probability ratio between old and new policies as follows:

\begin{equation}
    r(\theta) = \dfrac{\pi_\theta(a_t\vert o_t)}{\pi_{\theta_{old}}(a_t\vert o_t)}
\end{equation}

Then, the objective function of PPO (on policy) is the following:

\begin{multline}
    L^{CLIP}(\theta) = \mathbb{E}_{s,a} [\min\{r_t(\theta)\hat{A}(s, a),\\
    clip(r_t(\theta), 1-\epsilon, 1+\epsilon)\hat{A}(s, a)\}]
\end{multline}


where $\hat{A}$ is an estimate of the advantage function. PPO imposes its constraint by enforcing a small interval around 1, $[1-\epsilon, 1+\epsilon]$ to be exact, where $\epsilon$ is a hyperparameter. The function $clip(r(\theta), 1-\epsilon, 1+\epsilon)$ clips the ratio to be no more than $1+\epsilon$ and no less than $1-\epsilon$. The objective function of PPO takes the minimum of the original value and the clipped version, and thus we lose the motivation for increasing policy updates to extremes for better rewards. We use Generalized Advantage Estimation(GAE) \cite{Schulman2015} to estimate the advantage function $\hat{A}$. We adopt an open-source implementation of PPO \cite{pytorchrl2018} for our research.

\subsection{Frameworks: control architecture + learning method}
\begin{figure}[ht!] 
  \centering
     \begin{subfigure}[b]{0.45\textwidth}
         \centering
         \includegraphics[width=\textwidth]{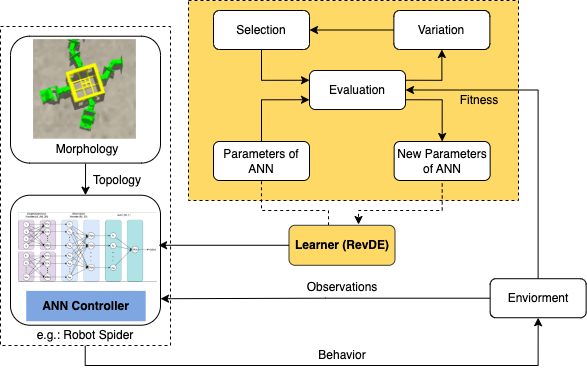}
         \caption{}
         \label{fig:a}
     \end{subfigure}
     \hfill
     \begin{subfigure}[b]{0.45\textwidth}
         \centering
         \includegraphics[width=\textwidth]{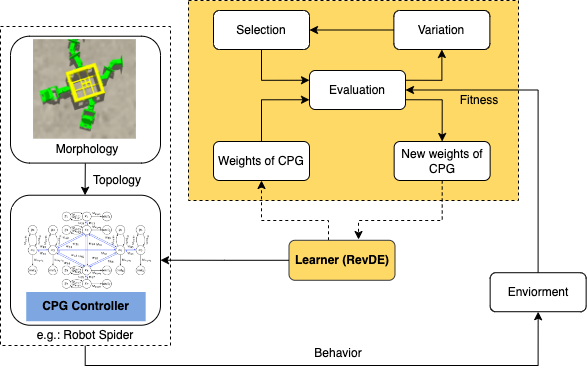}
         \caption{}
         \label{fig:b}
     \end{subfigure}
     \hfill
     \begin{subfigure}[b]{0.45\textwidth}
         \centering
         \includegraphics[width=\textwidth]{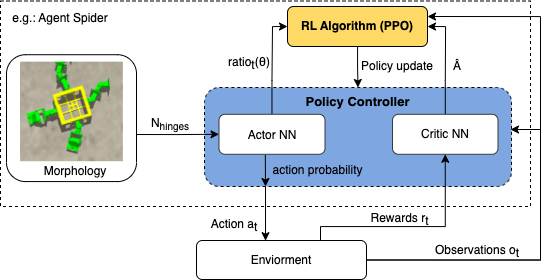}
         \caption{}
         \label{fig:b}
     \end{subfigure}
     \hfill
    \vspace{-0.8em}
  \caption{Schematic representations of three learning controller frameworks. The blue boxes show the controllers and the yellow boxes show the learners. In \ref{fig:framework}-(a), we show a spider as an example of robot morphology. The topology of the morphology determines the topology of the controller. The learner RevDE optimizes the weights of the CPG controller. In \ref{fig:framework}-(b), the learner RevDE optimizes the parameters of the ANN controller. \ref{fig:framework}-(c) is a DRL framework using PPO as the learning algorithm to change the parameters of two deep NNs to improve the policy controller for the tasks.}
  \label{fig:framework}
\end{figure}

We consider three combinations (frameworks) of control architectures and learning methods. The set-up of these three frameworks is shown in Figure \ref{fig:framework}. 

First, we use CPG-based controllers trained by a derivative-free method RevDE (see Figure \ref{fig:framework}-a). Second, we consider an MLP-based ANN controller trained by the same learner RevDE (see Figure \ref{fig:framework}-b). Lastly, we use a DRL-policy based controller trained by PPO (see Figure \ref{fig:framework}-c).

CPGs are not combined with PPO because our CPGs do not receive inputs (states). ANN and DRL are somewhat equivalent because they both are ANN-based controllers, however, DRL has one extra Critic NN therefore more parameters to be optimized. The outputs of the actor network in these two controllers are different too.

\vspace{-0.5em}

\subsection{Test suite of robot morphologies}

Given a set of robot modules and ways to attach them to functional robots, the number of possible configurations (thus, the number of possible robot morphologies) is, in general, infinite. For practically feasible empirical research, we need a limited set of robots to serve as test cases. In this paper, we use a test suite of twenty robots made of two parts: a set of viable and diverse robots produced by an evolutionary process, and a set of hand-picked robots \cite{Diggelen2021}. Regarding the first part, the key is to apply task-based fitness (viability) together with novelty search (diversity). The second part of the test suite can be filled by robots added manually by the experimenter. This option is entirely optional, it is to accommodate subjective preferences and interest in particular robot designs. The test suite we use here was generated by evolving a population of $500$ robots for speed and novelty w.r.t. the \textit{k-nearest neighbours} in the morphological space \cite{miras2018effects}. After termination, $15$ out of the $500$ robots were selected by maximizing the pairwise Euclidean distance in the morphology space. The other five robots (Gecko, Snake, Spider, BabyA, BabyB) were added manually. The robots are shown as inserts in Figure \ref{fig:fitness_mean_individual}.

\section{Experimental setup}
\subsubsection{Simulator}
We use a Mujoco simulator-based wrapper called Revolve2 to run the experiments. 
To have a fair comparison, we set the number of evaluations to be the same for each learner: 1000 learning evaluations. This number is based on the evaluations from RevDE for running 10 initial samples with 34 iterations. The first iteration contains 10 samples, and from the second iteration onwards each iteration creates 30 new candidates, resulting in a total of $10 + 30 \cdot (34-1)= 1000$ evaluations. Then with the same evaluation number 1000, we set PPO with 10 agents per iteration and 100 episodes. For the task of gait learning, we define the robot’s fitness as its average
speed in 30s, i.e. absolute distance in centimetres per second (cm/s).



\subsubsection{Setups and Code}
The code for carrying out the experiments is available online: \url{https://shorturl.at/gozS3}. A video showing examples of robots from the experiments can be found in \url{https://shorturl.at/gGHR3}. Table \ref{tab:experiments} shows the set-up of the experiments. The specific values of the hyperparameters are presented in Table \ref{tab:parameters}. 

\begin{table}[htp!]
\footnotesize
\caption{Experiments}
\centering
\begin{tabular}[b]{{p{0.22\linewidth}| p{0.4\linewidth}| p{0.18\linewidth}}}
\toprule
\textbf{Experiment}         & \textbf{Control Architecture} & \textbf{Learner} \\ \midrule
~CPG+RevDE 			 & ~CPG     & ~RevDE 	\\
~ANN+RevDE	        & ~ANN      & ~RevDE 	\\
~DRL+PPO		        & ~DRL-Policy      & ~PPO 	\\
\bottomrule 
\end{tabular}
\label{tab:experiments}
\end{table}

\begin{table}[htp!]
\footnotesize
\caption{Main experiment hyper parameters}
\centering
\begin{tabular}[b]{{p{0.32\linewidth}| p{0.09\linewidth}| p{0.48\linewidth}}}
\toprule
\textbf{CPG+RevDE}         & Value & Description \\ \midrule
~$\mu$ 			 & ~10     & ~Population size 	\\
~N 			 & ~30     & ~New candidates per iteration	\\
~$\lambda$ 			 & ~10     & ~Top-sample size 	\\
~$F$ 			 & ~0.5    & ~Scaling factor 	\\
~$CR$  & ~0.9    & ~Crossover probability \\ 
~Iterations  & ~34    & ~Number of iterations in RevDE \\ 

\bottomrule 
\toprule
\textbf{ANN+RevDE}         & Value & Description \\ \midrule
~$\mu$ 			 & ~30     & ~Population size 	\\
~N 			 & ~30     & ~New candidates per iteration	\\
~$\lambda$ 		& ~10     & ~Top-samples size 	\\
~$F$ 			 & ~0.5    & ~Scaling factor 	\\
~$CR$           & ~0.9    & ~Crossover probability \\ 
~Iterations  & ~34    & ~Number of iterations in RevDE \\ 

\bottomrule 
\toprule
\textbf{DRL+PPO}         & Value & Description \\ \midrule
~$\gamma$	   & ~0.2     & ~Discount gamma 	\\
~$\epsilon$	   & ~0.2     & ~PPO clipping parameter epsilon 	\\
~Entropy coefficient	   & ~0.01     & ~Entropy coefficient  	\\
~Value loss coefficient	 & ~0.5     & ~Value loss coefficient 	\\
~Episode		 & ~100     & ~A sequence of states, actions and rewards 	\\
~Agents			 & ~10     & ~Number of agents per episode 	\\
~Steps			 & ~150     & ~Number of steps before training 	\\
\bottomrule 
\end{tabular}
\label{tab:parameters}
\end{table}

\section{Results}
To compare the different frameworks, we consider three key performance indicators: efficiency, efficacy, and robustness to different morphologies.
\subsubsection{Efficacy}
The quality of a robot (fitness) is defined by the speed of the robot from the starting position to the stopping position within the simulation time.
The efficacy of a method is defined by the mean maximum fitness, averaged over the 20 independent repetitions: First, the maximum fitness achieved at the end of the learning process (1000 evaluations) is calculated within each independent repetition. Second, these maximum values are averaged over the 20 independent repetitions.

\begin{figure}[ht!] 
  \centering
  \includegraphics[width=0.95\linewidth]{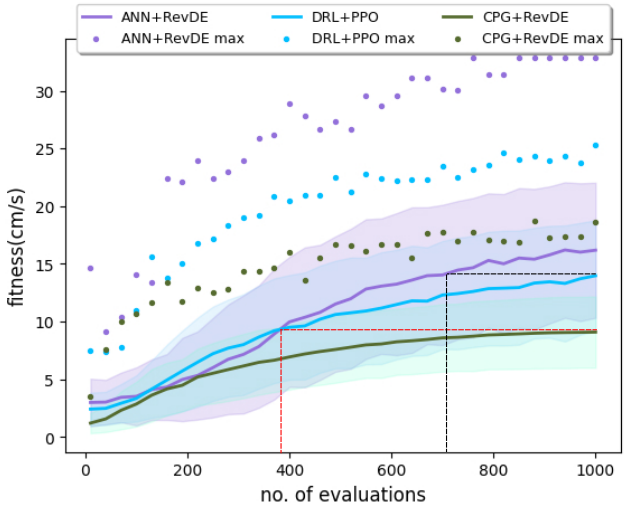}
  \caption{Mean fitness over 1000 evaluations across morphologies (averaged over 20 runs) for 3 experiments. The dots indicate the mean maximum fitness in each evaluation (averaged over 20 runs). The shaded areas show the standard deviation.}
  \label{fig:fitness_mean_max}
\end{figure}

In Figure \ref{fig:fitness_mean_max}, the dots indicate the maximum fitness in each evaluation (averaged over 20 runs). We can see that with the same learner (RevDE), the ANN controller outperforms the CPG controller significantly. ANN+RevDE achieves a two times higher fitness value compared to CPG+RevDE at the end of the 1000 evaluations. This could be due to CPGs producing more connected actions with fewer controller parameters, while ANNs have much more parameters to optimize and output the action probability instead of the action itself which produces actions that are very different even in subsequent time steps, eventually helping the exploration. The mean maximum fitnesses of ANN+RevDE and DRL+PPO have no significant difference initially but after evaluation 200, ANN+RevDE yields much higher fitness values than DRL+PPO. 

\begin{figure}[htbp] 
  \centering
  \includegraphics[width=0.90\linewidth]{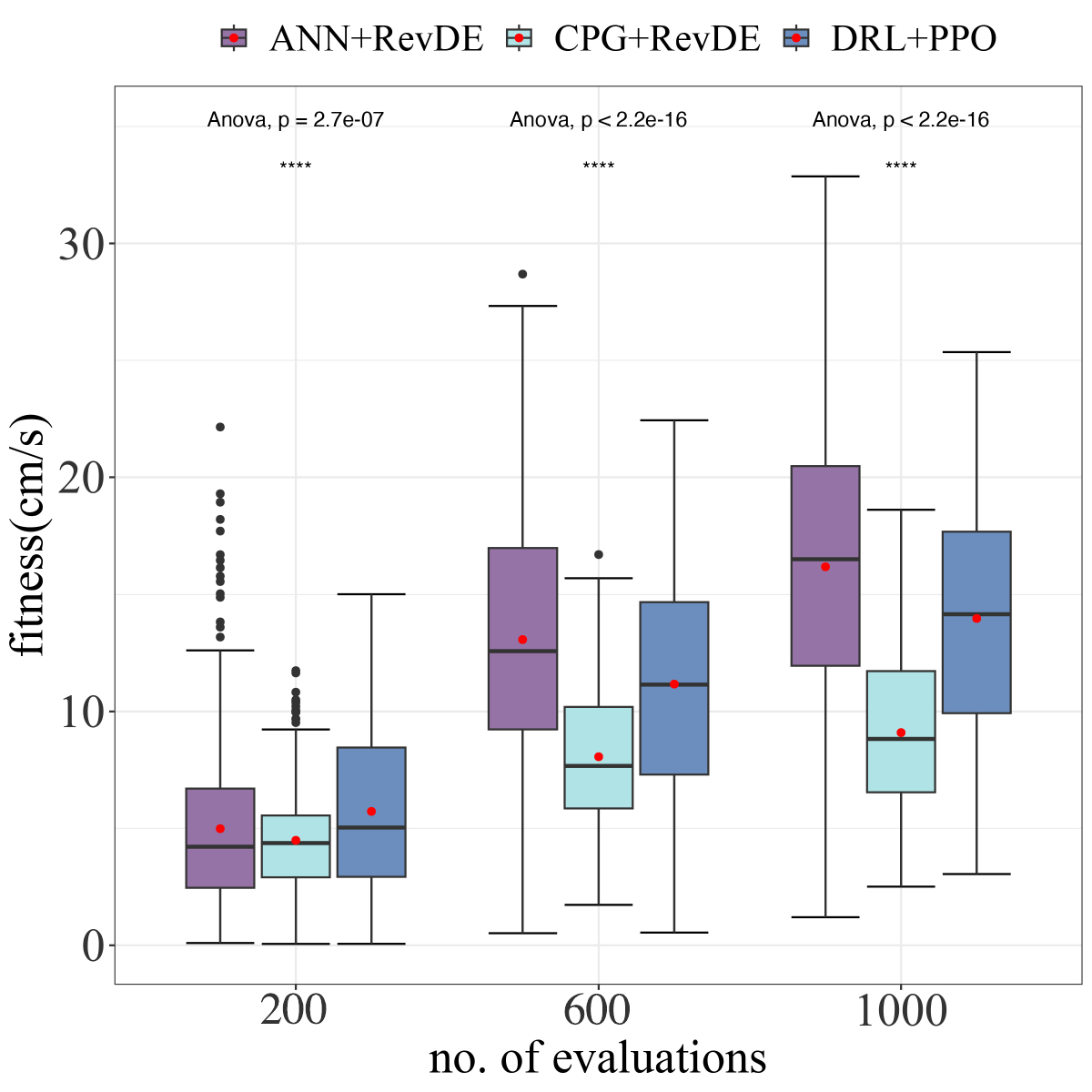}
  \caption{Efficacy boxplot. Validation of three frameworks at three evaluations. Red dots show mean values. }
  \label{fig:framework_3evaluations}
\end{figure}

Second, another way to measure the quality of the solution is by giving the same computational budget (number of evaluations) and measuring which method finds the best solution (highest fitness) faster. In Figure \ref{fig:framework_3evaluations}, it is more significantly different among these three frameworks at evaluations 600 and 1000 than 200. Given the evaluations of 200, DRL+PPO has the highest mean fitness value. While at the evaluations of 600 and 1000, ANN+RevDE surpasses DRL+PPO.  With regards to CPG+PPO, the fitness increasing speed is slower than the other two methods. 

\begin{figure}[ht!] 
  \centering
     \begin{subfigure}[b]{0.47\textwidth}
         \centering
         \includegraphics[width=\textwidth]{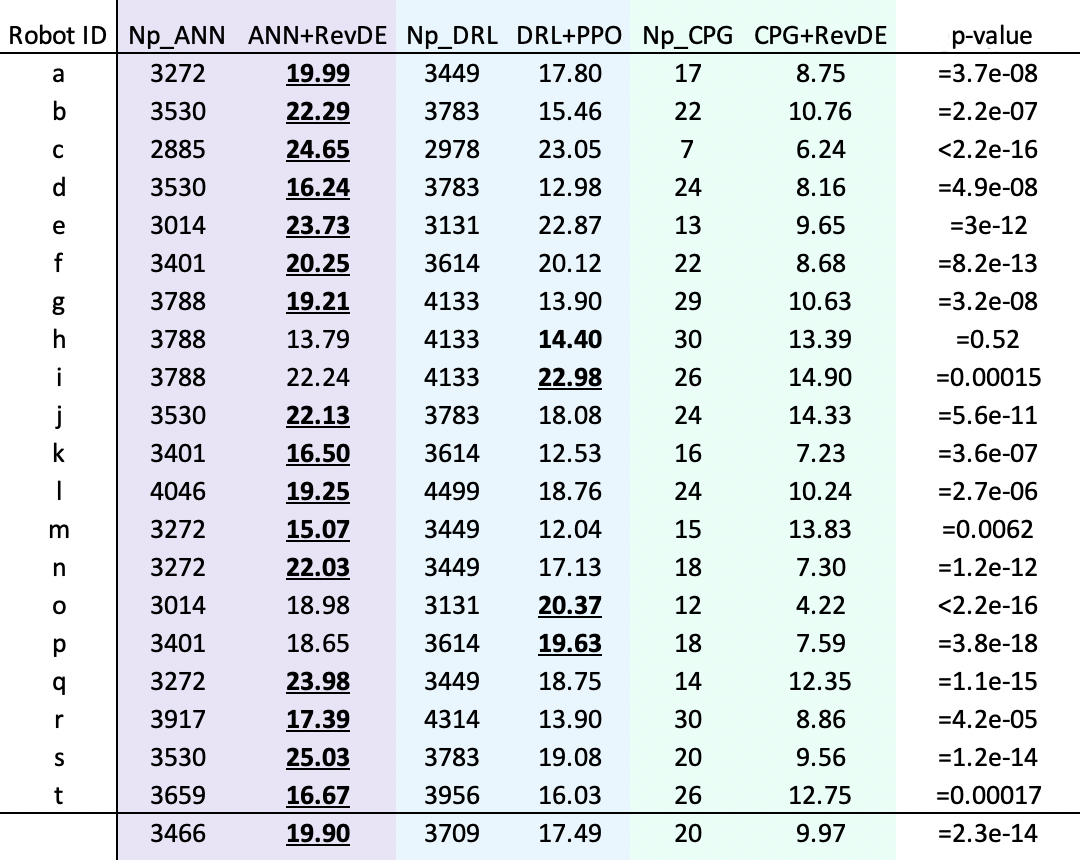}
         \caption{}
         \label{fig:a}
     \end{subfigure}
     \hfill
     \begin{subfigure}[b]{0.47\textwidth}
         \centering
         \includegraphics[width=\textwidth]{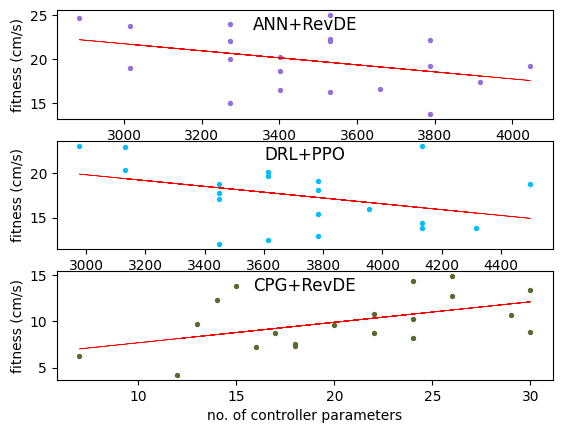}
         \caption{}
         \label{fig:b}
     \end{subfigure}
     \hfill
    \vspace{-0.8em}
  \caption{(a) Mean maximum fitness per morphology for each framework. For each robot morphology, there're columns of controller parameters numbers, namely Np\_ANN in purple, Np\_DRL in blue and Np\_CPG in green and the best result which is indicated with boldface while underline indicates significantly better performance compared to the other frameworks. The last row shows the aggregated result for each framework over all morphologies. (b) the correlation between the number of controller parameters and the mean maximum fitness per framework. The red lines are the linear regression lines.}
  \label{fig:fitness_MBF_params} \vspace{-1em}
\end{figure}

\subsubsection{Efficiency}
Efficiency indicates how much effort is needed to reach a given quality threshold (the fitness level): it is measured as the average number of evaluations to `find a solution'. Figure \ref{fig:fitness_mean_max} displays the usual quality-versus-effort plots, specifically the mean fitness over the number of evaluations. Looking at the solid curves reveals that ANN+RevDE is more efficient than the other methods. As marked by the red dotted lines, it takes only 370 evaluations for ANN+RevDE (purple curve) to reach the level of fitness that the CPG+RevDE method (green curve) achieves at the end of the learning period, 1000 evaluations. Similarly, the black dotted lines mark the number of evaluations at 730 when ANN+RevDE achieved the levels of fitness that DRL+PPO reached after 1000 evaluations.

\subsubsection{Robustness} 
The robustness of a framework is defined by the variance in different robot morphologies. We can measure this by the variance of a framework’s mean maximum fitness over the robot zoo and the mean fitness per robot over the number of evaluations. 

\begin{figure*}[ht!]
\input{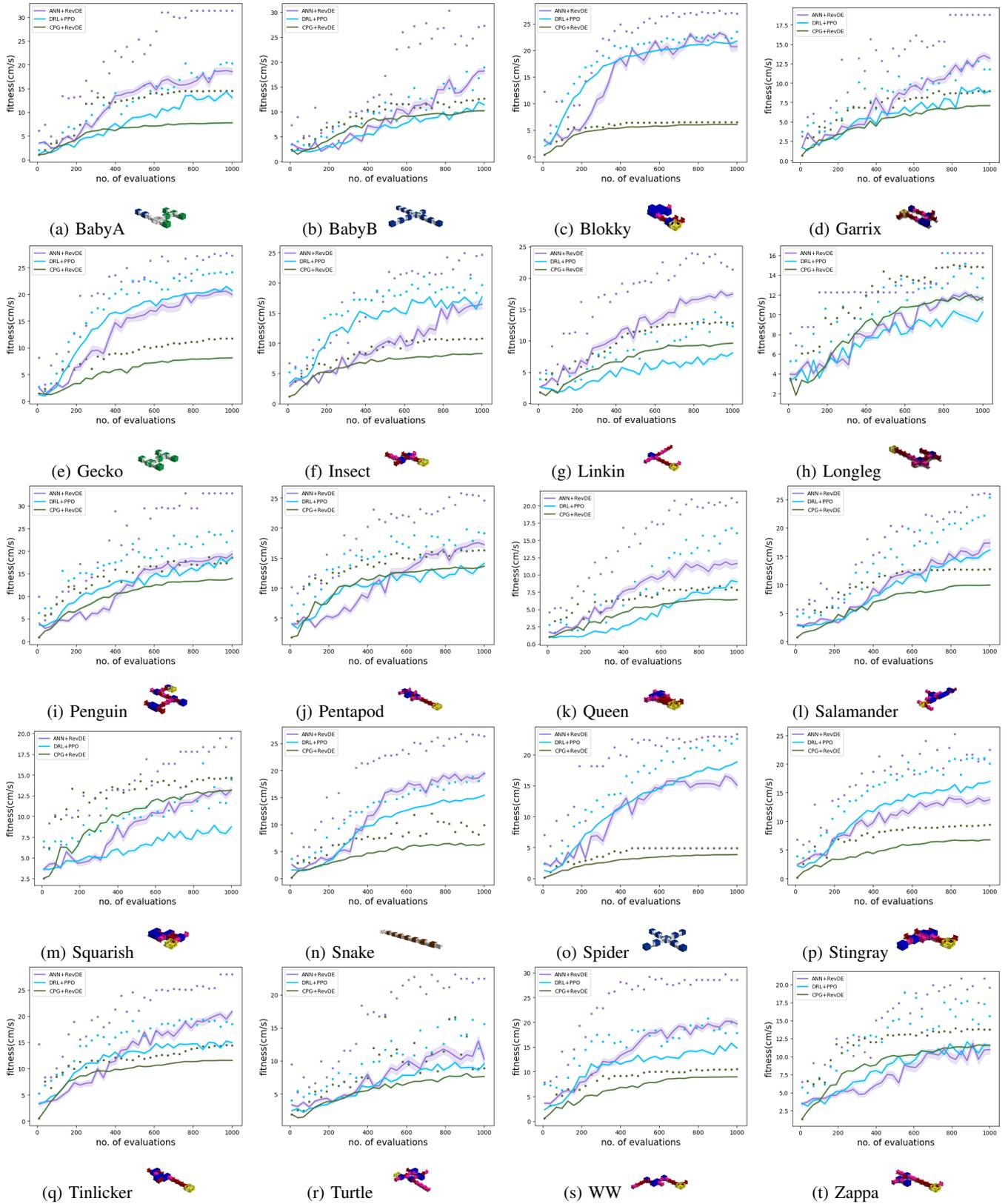}
\vspace{-0.5 em}
\caption{Mean speed over time (the number of evaluations) during gait-learning, averaged over 20 independent repetitions per robot. The purple lines show the ANN+RevDE, the blue lines indicate DRL+PPO and the green lines indicate CPG+RevDE. The bands indicate the 95\% confidence intervals ($\pm1.96\times SE$, Standard Error).}
\label{fig:fitness_mean_individual} 
\end{figure*}

Figure \ref{fig:fitness_mean_individual} shows the mean and maximum fitness of three frameworks over the number of evaluations per robot from Robot Zoo. The bands indicate the 95\% confidence intervals ($\pm1.96\times SE$, Standard Error). CPG+RevDE has a narrower band than the other two frameworks which indicates lower uncertainty and is more stable. In Figure \ref{fig:fitness_MBF_params}-a, we present a numerical summary of the results of the mean maximum fitness per robot per framework. It shows ANN+RevDE outperform CPG+RevDE or DRL+PPO on 16 robots significantly: a, b, c, d, e, f, g, j, k, l, m, n, q, r, s, t. DRL+PPO wins on robot i, o, p significantly and on robot h non-significantly. 

Figure \ref{fig:fitness_MBF_params}-b exhibits the correlation between the number of controller parameters and mean maximum fitness per framework. The dots in each plot represent 20 robot zoo. The results indicate that the deep neural network-based frameworks (ANN+RevDE and DRL+PPO) show a negative linear relationship between the number of controller parameters and the fitness value while the CPG-based framework shows a positive linear relationship. Among the frameworks, it shows that the significant difference in the number of controller parameters between the Deep NN-based and the CPG-based controllers does reflect on their fitnesses (different y-scales).







\section{Conclusions and Future Work}

This work investigated different combinations of control architectures with learning algorithms applied to a diverse set of robot morphologies.

Regarding efficacy and efficiency, the ANN+RevDE framework achieved levels of quality that the other two frameworks managed to achieve only at later stages of the learning period. As for robustness, all three frameworks successfully optimized all robots. However, ANN+RevDE outperformed DRL+PPO or CPG+RevDE significantly on 16 robots, while DRL+PPO outperformed on only 3 robots. Therefore ANN+RevDE is the best-performing learning controller framework in all three measures.

Interestingly, with the same learning algorithm (RevDE), the CPG controller performs more steadily with a lower standard deviation while the ANN controller takes longer to explore at the beginning, then it increases steeply with a much higher standard deviation (Figure \ref{fig:fitness_mean_max} and \ref{fig:framework_3evaluations}). This can be due to the significant difference in the number of parameters in different controllers, but future research is needed to investigate this phenomenon. 




\bibliographystyle{IEEEtran}
\bibliography{references} 
\end{document}